%% file: main.tex
\documentclass{article}

\PassOptionsToPackage{numbers, compress}{natbib}

    \usepackage[preprint]{neurips_2023}

\usepackage[utf8]{inputenc} 
\usepackage[T1]{fontenc}    
\usepackage{hyperref}       
\usepackage{url}            
\usepackage{booktabs}       
\usepackage{amsfonts}       
\usepackage{nicefrac}       
\usepackage{microtype}      
\usepackage{xcolor}         
\usepackage{amsmath}
\usepackage{multirow}
\usepackage{array}
\usepackage{textcomp}
\usepackage{graphicx}
\usepackage{multicol}
\usepackage{multirow}
\usepackage{wrapfig}
\usepackage{CJKutf8}
\usepackage[english]{babel}
\usepackage{subfig}
\usepackage{pifont}
\usepackage{float}
\usepackage{xspace}
\usepackage{array}
\usepackage{array}
\usepackage{marvosym}
\usepackage{tablefootnote}
\newcolumntype{L}[1]{>{\raggedright\let\newline\\\arraybackslash\hspace{0pt}}m{#1}}
\newcolumntype{C}[1]{>{\centering\let\newline\\\arraybackslash\hspace{0pt}}m{#1}}
\newcolumntype{R}[1]{>{\raggedleft\let\newline\\\arraybackslash\hspace{0pt}}m{#1}}
\newcommand{\Ours}{Zero-shot Temporal Aligner\xspace}
\newcommand{\ours}{ZeroTA\xspace}

\title{Zero-Shot Dense Video Captioning by \\ Jointly Optimizing Text and Moment}

\author{%
  Yongrae Jo \\
  KAIST AI\\
  \texttt{yongrae@kaist.ac.kr} \\
  \And
  Seongyun Lee \\
  Korea University \\
  \texttt{sy-lee@korea.ac.kr} \\
  \AND
  Aiden SJ Lee\\
  Twelve Labs \\
  \texttt{aiden@twelvelabs.io} \\
  \And
  Hyunji Lee \\
  KAIST AI \\
  \texttt{alee6868@kaist.ac.kr} \\
  \And
  Hanseok Oh \\
  KAIST AI \\
  \texttt{hanseok@kaist.ac.kr} \\
  \And
  Minjoon Seo \\
  KAIST AI \\
  \texttt{minjoon@kaist.ac.kr} \\
}

\begin{document}

\maketitle

\begin{abstract}

Dense video captioning, a task of localizing meaningful moments and generating relevant captions for videos, often requires a large, expensive corpus of annotated video segments paired with text.
In an effort to minimize the annotation cost, we propose \ours, a novel method for dense video captioning in a \textit{zero-shot manner}. 
Our method does not require any videos or annotations for training; instead, it localizes and describes events within each input video at test time by optimizing solely on the input.
This is accomplished by introducing a soft moment mask that represents a temporal segment in the video and jointly optimizing it with the prefix parameters of a language model.
This joint optimization aligns a frozen language generation model (i.e., GPT-2) with a frozen vision-language contrastive model (i.e., CLIP) by maximizing the matching score between the generated text and a moment within the video.
We also introduce a pairwise temporal IoU loss to let a set of soft moment masks capture multiple distinct events within the video.
Our method effectively discovers diverse significant events within the video, with the resulting captions appropriately describing these events.
The empirical results demonstrate that \ours surpasses zero-shot baselines and even outperforms the state-of-the-art few-shot method on the widely-used benchmark ActivityNet Captions. Moreover, our method shows greater robustness compared to supervised methods when evaluated in out-of-domain scenarios.
This research provides insight into the potential of aligning widely-used models, such as language generation models and vision-language models, to unlock a new capability—understanding temporal aspects of videos.

\end{abstract}

\section{Introduction}

\begin{figure}[h]
\centering
\includegraphics[width=\columnwidth]{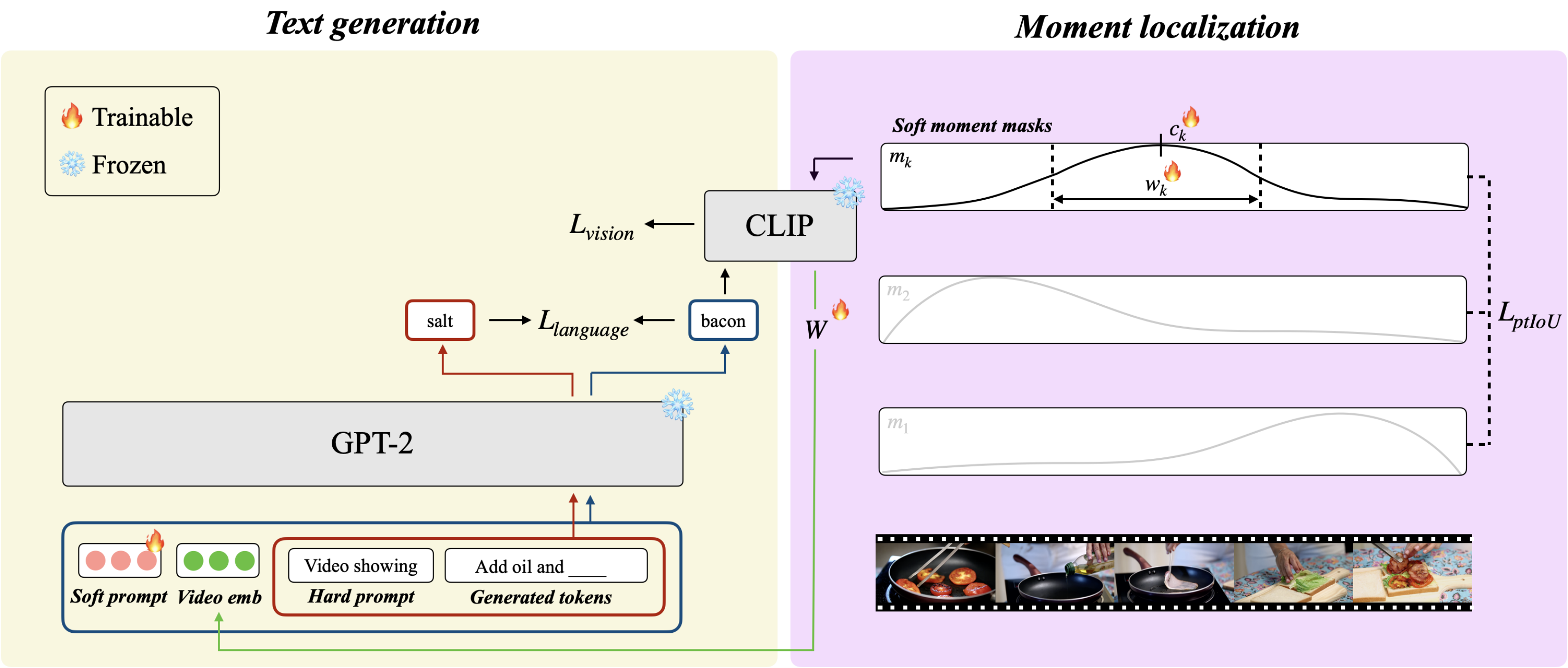}
\caption{Architecture of \ours and training connections. \ours consists of two modules: text generation and moment localization. Text generation is conditioned on the concatenation of a soft prompt, projected video embedding, and hard prompt. Among these, the soft prompt and the video embedding projection layer (\textbf{W}) are trainable. Temporal localization is accomplished by soft moment masks parameterized with trainable center ($c_k$) and width ($w_k$) parameters. There are three losses in our model: vision loss~($L_{vision}$), language loss~($L_{language}$), and pairwise temporal IoU loss~($L_{ptIoU}$). $L_{vision}$ measures a matching score between the current generated token and the visual information using CLIP. $L_{language}$ is computed between token probability distributions produced with (blue box input) and without (red box input) the trainable prefix. The $L_{ptIoU}$ measures how much overlapped the moments are and make them apart from each other. All trainable parameters are optimized only on a single input video at test time.}
\label{fig:method}
\end{figure}

Dense video captioning is a task that temporally localizes multiple meaningful events (or moments) within a video and provides captions for each event \citep{krishna2017dense, wang2021end}.
As \emph{wild} videos are often untrimmed and contain multiple events within a single video, this task is particularly useful in real-world scenarios.
Dense video captioning requires a deep understanding and accurate representation of temporal information present in the video. As a result, it typically requires a substantial collection of annotations for temporal segments within videos, each paired with corresponding captions, which is often prohibitively costly.

For this reason, performing dense video captioning without access to language captions or annotated temporal segments is especially valuable, but the literature lacks previous work on a zero-shot setup.
In this paper, we propose \ours (\Ours), which tackles the problem of zero-shot dense video captioning by jointly optimizing text generation and moment localization for a single video at test time in an end-to-end manner. 
This joint optimization ensures that the generated text aligns with the discovered temporal moment and, simultaneously, that the discovered temporal moment accurately corresponds to the generated text.

Our model comprises two modules: the text generation module and the moment localization module (Figure~\ref{fig:method}). The design of the text generation module is inspired by \citep{tewel2022zerocap, tewel2022zero}, where they address image and video captioning tasks without training data. Likewise, we leverage a frozen language generation model (i.e., GPT-2~\citep{radford2019language}) and a frozen vision-language model (i.e., CLIP~\citep{radford2021learning}), and align GPT2 with CLIP using a small number of learnable prefix parameters as in prefix-tuning \citep{li2021prefix}.
Although CLIP is pretrained on image-text pairs with contrastive learning and GPT-2 is pretrained on text-only data without video knowledge, \ours can effectively localize and generate captions for different moments in a video. This work hints at how we can align models such as a language generation model and a vision-language contrastive model, to build a compositional model that is capable of temporal understanding.

For the design of the moment localization module, we propose one new masking mechanism and one new loss term: soft moment masking and pairwise temporal IoU loss. The soft moment masking ensures the text generation focuses solely on the corresponding video moment by introducing a differentiable temporal mask onto video frames. Pairwise temporal Intersection over Union (IoU) loss ensures that our approach generates multiple captions from distinct time segments within a video, thus enhancing the richness of the dense captions. This loss is calculated on a group of moments that are jointly optimized for a given video.

We validate the effectiveness of our approach in accurately identifying and describing significant moments in a given video.
Our zero-shot method surpasses various zero-shot baselines and outperforms the state-of-the-art few-shot method pretrained on a billion-scale video-text data \citep{yang2023vid2seq} on the widely-used ActivityNet Captions benchmark.
Furthermore, we demonstrate the robustness of our method in out-of-domain scenarios when compared to supervised models. When assessed on a dataset distinct from the one used for model training, supervised models struggle to adapt to the new dataset. Conversely, our zero-shot approach exhibits better resilience in this situation. The out-of-domain setup is especially valuable when seeking to make use of real-world videos, which are characterized by distinctly different domains.

To summarize, we provide the following key contributions: 
\begin{itemize}
    \item We propose \ours (\Ours), a pioneering zero-shot dense video captioning method by aligning pretrained models to unlock a new capability of temporal understanding.
    \item We propose soft moment masking for end-to-end optimization of temporal localization and pairwise temporal IoU loss for the diversity of localized moments.
    \item Our method surpasses various zero-shot baselines and even outperforms the state-of-the-art few-shot method on the ActivityNet Captions benchmark. Also, our method is more robust in out-of-domain scenarios than supervised models.
\end{itemize}

\section{Related Work}
\label{related_work}

\paragraph{Dense video captioning}

Dense video captioning (also called dense event captioning \citep{krishna2017dense}) extends the task of video captioning \citep{gao2017video, lin2022swinbert, pan2017video, wang2018reconstruction, wang2018video} by incorporating fine-grained temporal localization and generate multiple captions per video.
Due to the complexity of the task, most existing methods \citep{krishna2017dense, iashin2020better, iashin2020multi, wang2018bidirectional, wang2020event, wang2021end, deng2021sketch, zhang2022unifying, zhu2022end} require a strong supervision with a large amount of video-text-timestamp data.

To mitigate annotation costs, existing attempts \citep{duan2018weakly, chen2021towards, rahman2019watch} have focused on addressing dense video captioning tasks with lower levels of supervision.
Specifically, \citet{duan2018weakly} introduced a weakly supervised methodology for dense video captioning, utilizing video data paired with captions without time intervals annotation during the training process. However, these approaches still rely on video paired with text corpus and make a somewhat unrealistic assumption of a one-to-one correspondence between video segments and their respective captions. In contrast, we present a zero-supervision paradigm that eliminates the need for a video or text corpus for training.

\citet{yang2023vid2seq} recently introduced a few-shot dense video captioning setup, which involves first pretraining a model on narrative videos and then fine-tuning it with a small portion of the downstream training data. Our approach extends this few-shot setting further by introducing zero-shot dense video captioning. Also, our method does not need pretraining on video data.

\paragraph{Vision-language alignment}

Our approach is related to the models that bridge between visual and textual modalities. CLIP~\citep{radford2021learning} is one such model that has gained noteworthy recognition. Recent works \citep{merullo2022linearly, liu2023visual, tsimpoukelli2021multimodal, eichenberg2021magma, alayrac2022flamingo} show that pretrained image and text models can be tuned together to be applied to various vision-language tasks.

In particular, \citet{merullo2022linearly} showed visual representations from frozen vision models can be projected onto frozen language models with a single linear layer. Similarly, \citet{liu2023visual} connected image features into the word embedding space using a trainable projection matrix. Our method follows a similar approach and incorporates projected visual embeddings as a prefix into a frozen language model.

\citet{tewel2022zerocap, tewel2022zero} combine a visual-semantic model with a language model, leveraging knowledge from both models to generate descriptive text given an image or a video, respectively. Inspired by these works, we take a step further to apply this approach to solve zero-shot dense video captioning tasks for the first time. Notably, The task of dense video captioning requires a temporal understanding of a video, which an image-text visual-semantic model has never been trained on.

\paragraph{Moment localization}

Moment localization is the task of identifying specific moments from a video that are relevant to a given natural language query \citep{chen2019localizing, gao2017tall, lu2019debug, zeng2020dense, zhang2019man, mun2020local, rodriguez2021dori, rodriguez2020proposal}.
Since obtaining annotations for moment localization can be costly, several studies have explored ways to lessen the need for supervision. As part of these efforts, the weakly supervised setup for moment localization has been proposed \citep{gao2019wslln, mithun2019weakly, ma2020vlanet, wang2021visual, yoon2021weakly}. Although these methods reduce the costs related to temporal annotations, the remaining cost associated with the creation of natural language queries continues to be significant.

A few works explored zero-shot setup for moment localization \citep{nam2021zero, wang2022prompt, jiang2022pseudo, kim2023language, paul2022text}. \citet{nam2021zero} extract nouns and verbs from moment proposals by object detection and simple language modeling, then use them as pseudo-queries to train a moment localization model. While this method produces simplified sentences resembling dense video captions during the procedure, the constructed queries are mere lists of nouns and verbs that lack natural language properties. As such, they are not designed to address the dense video captioning task. Similarly, \citet{kim2023language} takes a simpler approach to zero-shot moment localization by utilizing CLIP, but it does not generate discrete captions in natural language.

\section{Method}
\label{method}

\begin{figure}[h]
\centering
\includegraphics[width=120mm]{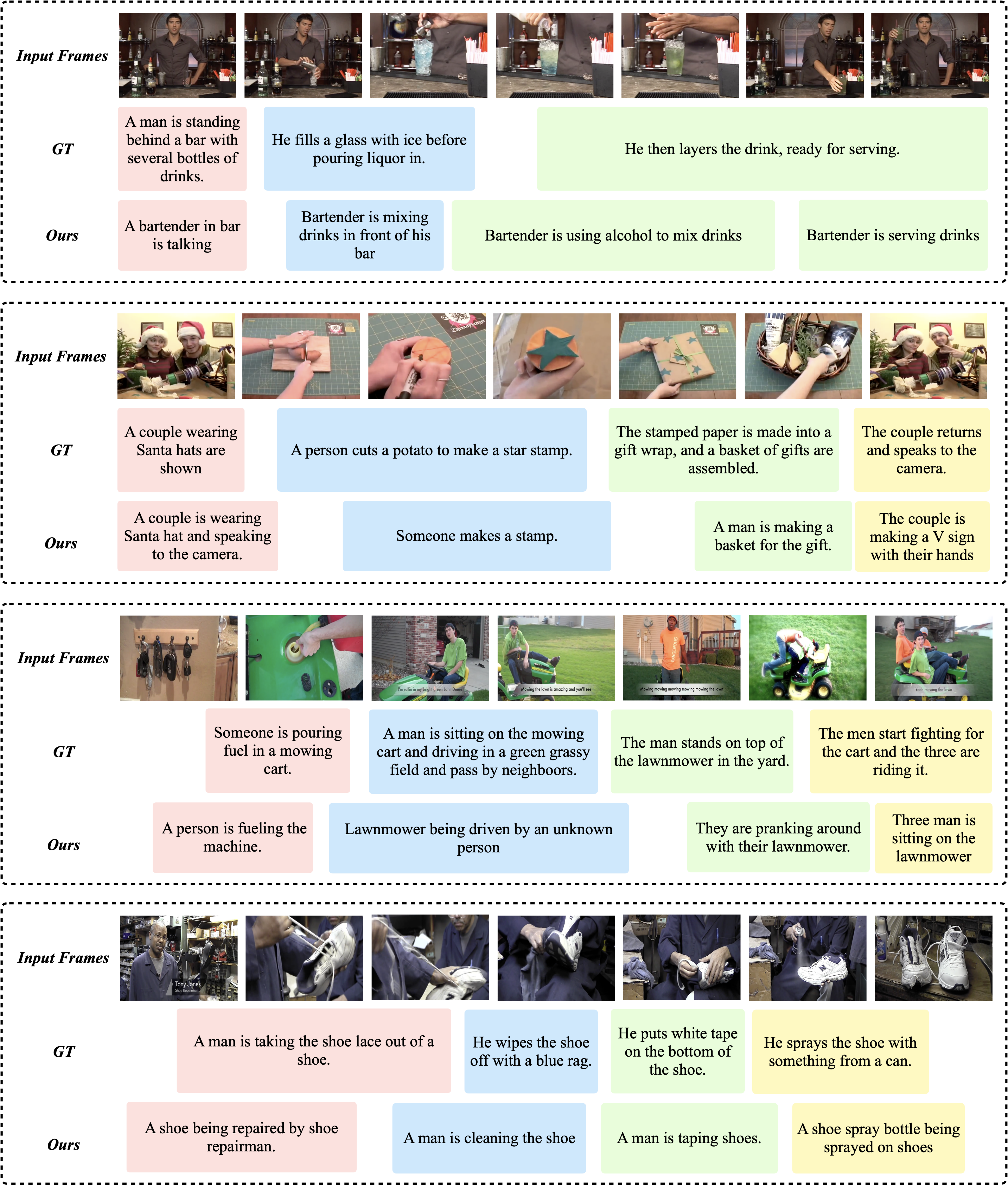}
\caption{Example of dense video captioning predictions of \ours on ActivityNet Captions validation set, compared with ground-truth.}
\label{fig:example}
\end{figure}

Dense video captioning aims to describe with natural language events within a given untrimmed video, while also temporally localizing them with start and end time stamps (Figure~\ref{fig:example}).
Formally, the task of dense video captioning can be described as follows: Given a video $\mathbf{V}$ of $L$ frames, the objective is to determine a function $F: \mathbf{V} \rightarrow \{(s_k, m_k)\}^{N}_{k=1}$ where $s_k$ represents the caption, $m_k$ denotes the corresponding moment, and $N$ is the number of moments.
Each caption $s_k$ is a sequence of tokens, and each moment $m_k$ is a consecutive subset of video frames. A moment signifies a meaningful temporal segment of the video. In this work, we treat $N$ as a hyperparameter that is predetermined before the input is given.

In zero-shot dense video captioning, the model does not have access to language captions or annotated time stamps for training.
Therefore, the challenges are two-fold.
First, the model needs to accurately identify significant moments within a long video without annotated captions.
Second, it must generate natural language captions for each of these identified moments, without annotated moments.

To tackle these two challenges at the same time, we design a training-free method, \ours~(\Ours). As in Figure~\ref{fig:method}, \ours is composed of two modules. The first is the text generation module (left of Figure~\ref{fig:method}) that utilizes a frozen language model, which is conditioned on a learnable prefix context. The prefix context and vision loss~($L_{\text{vision}}$) are designed to produce text that aligns with the visual content corresponding to a specific moment, as detailed in Section~\ref{method:text_generation}. The second module, referred to as the moment localization module (right part of Figure~\ref{fig:method}), is responsible for learning the parameters that specify a moment in a video while ensuring the diversity of moments, as presented in Section~\ref{method:moment_localization}. Finally, we combine the losses for both modules and optimize the model in an end-to-end manner, as described in Section~\ref{method:joint_optimization}.

\subsection{Text generation}
\label{method:text_generation}
The text generation module uses a pretrained language model (i.e., GPT-2) to infer the next word from a prefix context.
The language model parameters are fixed, and only the prefix context parameters~(Section~\ref{method:prefix_context}) are optimized during the test time to align the generated text to the corresponding moment.
The optimization takes place during auto-regression and is iterated for each generation step.

Taking inspiration from using a vision-language alignment model and a language model for image and video captioning~\citep{tewel2022zerocap, tewel2022zero}, we adopt two losses during the optimization process for the text generation module. The first loss, the vision loss~(Section~\ref{method:vision_loss}), aims to enhance the similarity between the generated text and the corresponding moment. The second loss, the language loss~(Section~\ref{method:lang_loss}), focuses on preserving the naturalness of the generated text.

\subsubsection{Prefix context}
\label{method:prefix_context}

The prefix context has three parts: soft prompt, projected video embedding, and hard prompt.
These three parts are concatenated and used by the language model as a prefix for language generation.

The first part of the prefix context is the tunable soft prompt. 
Similar to the prefix-tuning \citep{li2021prefix}, the transformer blocks within the length of the soft prompt have their key and value embeddings learned during the optimization process.
The frozen language model then attends to this soft prompt, providing guidance during the generation process.

The second part of the prefix context is the projected video embedding.
To obtain these embeddings, we first extract the image features from video frames using a pretrained image encoder (CLIP image encoder), aggregate the features with a soft moment mask (Section~\ref{method:moment_localization}), and then apply a simple trainable projection layer ($\textbf{W}$) to the aggregated video feature embedding.
The trainable projection layer is a single linear layer used to project video feature embedding to language model token embedding space \citep{merullo2022linearly}. By the projection, the dimensionality of video feature embedding matches that of the language model token embedding.

The third part of the prefix context is the hard prompt.
These are prefix tokens such as 'Video showing,' 'Video of,' etc. We randomly sample a hard prompt from a list of prefix tokens. The list of the prefix tokens we used in experiments is in the Appendix Section \ref{sec:appendix:imple}.

\subsubsection{Vision loss}
\label{method:vision_loss}

To steer the language model toward a specific visual direction at each generation step, we incorporate vision loss.
This loss is obtained through a vision-language alignment model (CLIP). CLIP scores the relevance between the generated tokens up to the current step and a video moment, which we call the alignment score (Eq.~\ref{eq:vision}). For $i$-th candidate token $t^i_{k,l}$ at generation step $l$ of caption $s_k$, we form the associated candidate sentence $s^i_{k,l}$ by concatenating the candidate token with previously generated tokens $s^i_{k,l} = \{t_{k,1}, \ldots, t_{k,l-1}, t^i_{k,l}\}$ and calculate alignment score for each candidate sentence\footnote{For efficiency, we compute the scores only for the top 512 candidate tokens.}.

The alignment score ($a^i_{k,l}$) of the $i$-th candidate token at generation step $l$ of caption $s_k$ is computed as

\begin{equation}
\begin{aligned} \label{eq:vision}
    a^i_{k,l} \propto \exp(\cos(\text{E}_{\text{Text}}(s^i_{k,l}), \text{E}_{\text{Image}}(m_k)) / \tau)
\end{aligned}
\end{equation}

where $\cos$ denotes the cosine similarity, and $\text{E}_{\text{Text}}$ and $\text{E}_{\text{Image}}$ represent the textual and image encoder of the vision-language alignment model (CLIP). This measures the similarity between the textual embedding of candidate sentence $s^i_{k,l}$ and the image embedding of the moment $m_k$. $\tau > 0$ is a temperature hyperparameter.

The vision loss is defined as the average cross-entropy loss ($CE$) between the alignment score distribution ($a_{k,l}$) and the probability distribution of the candidate tokens ($q_{k,l}$) obtained by the language model:

\[
    L_{\text{vision}} = \frac{1}{N}\sum_k CE(a_{k,l}, q_{k,l})
\]

This loss stimulates token generation towards higher text-visual matching scores between the generated text and visual information from the moment.

\subsubsection{Language loss}
\label{method:lang_loss}

In order to preserve the natural language quality of the generated text while aligning it with the visual content, we employ a regularization term, which we call language loss.
This loss quantifies the average cross-entropy ($CE$) between the probability distribution of words from the language model with the prefix context ($q_{k,l}$) and without the prefix context ($q'_{k,l}$). 
By minimizing this loss, we ensure that the probability distribution of words with the prefix context closely matches that of the original language model without the prefix context.
This regularization step helps maintain the overall language model coherence while incorporating visual alignment \citep{tewel2022zerocap, tewel2022zero}.

\[ L_{\text{language}} = \frac{1}{N}\sum_k CE(q_{k,l}, q'_{k,l}) \]

\subsection{Moment localization}
\label{method:moment_localization}

Similar to how the text generation module aligns generated text with a video moment, the moment localization module is responsible for aligning the video moment with the generated text.
Previous works performed the selection of temporal moments through a separate module, relying solely on visual feature similarity \citep{nam2021zero, kim2023language}.
However, such an approach is sub-optimal as the moments are selected without considering the corresponding captions.
To remedy this, we introduce soft moment masking~(Section~\ref{method:soft_mask}).

Dense video captioning requires identifying multiple temporal moments from a given video. To accomplish this, instead of generating a single moment-text pair, we optimize a group of moments simultaneously. In order to enhance the diversity among temporal moments and ensure that each moment captures distinct meaningful segments of a video, we introduce the pairwise temporal IoU loss~(Section~\ref{method:ptIou_loss}).

\begin{figure}[h]
\centering
\includegraphics[width=80mm]{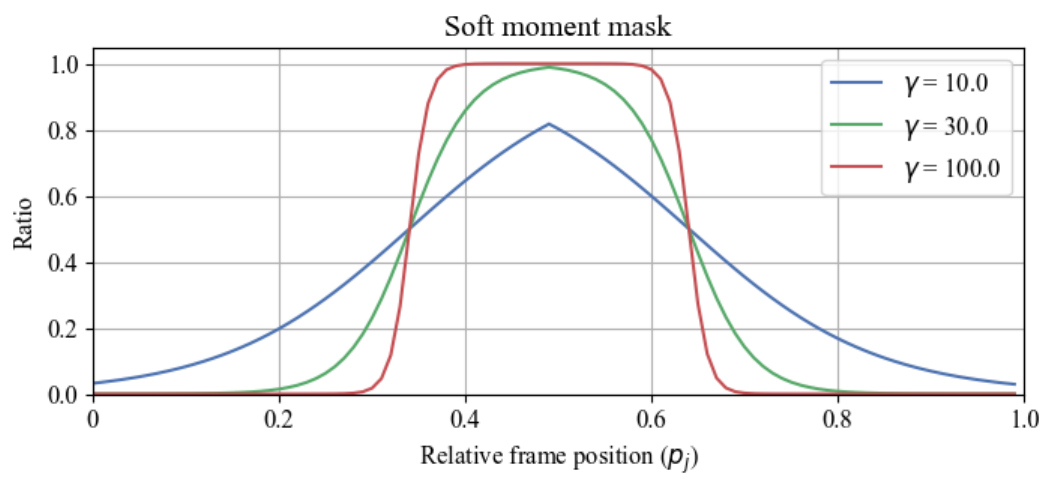}
\caption{Visualization of the soft moment mask under varying sharpness hyperparameters~($\gamma$), while keeping the center and width fixed. The relative frame position ($p_j$) denotes the normalized position of a frame within the video length, with 0 indicating the video's start and 1 indicating its end. As the sharpness increases, the contrast between the mask ratio inside and outside the moment also increases. We gradually increase the sharpness through optimization iterations.}
\label{fig:sharpness}
\end{figure}

\subsubsection{Soft moment masking}
\label{method:soft_mask}

A soft moment mask specifies a moment $m_k$ with two parameters: center $c_k$ and width $w_k$.
These two parameters are randomly initialized and tuned during end-to-end optimization.
To construct a soft mask that spans the length of the video using the two parameters, we employ the following steps:

\begin{enumerate}
  \item Apply the sigmoid function to $c_k$ and $w_k$ to convert it to normalized values $0 \leq \tilde{c}_k \leq 1$ and $0 \leq \tilde{w}_k \leq 1$ that indicate their relative positions to the length of the video; 0 represents the start of the video, and 1 represents the end of the video.
  \item Let $p_j \in \{p_1, \ldots, p_L\}$ denote the normalized frame position value between 0 and 1, representing the relative position of a frame to the length of the video.
  \item Calculate the L1 distance between each frame position $p_j$ and $\tilde{c}_k$.
  \item Subtract an offset of half the normalized width ($\tilde{w}/2$) from the distance, multiply it by the sharpness hyperparameter $\gamma$, and then apply the sigmoid function on it.
    
\end{enumerate}

We can summarize the above procedure with the following formula. The value of $j$th position in the mask for the moment $m$, $\text{mask}_{j}$, is

\[
  \text{mask}_{j} = \text{sigmoid}(\gamma * (|p_j - \tilde{c}_k| - \tilde{w}_k / 2)) \text{ where } \tilde{c}_{k} = \text{sigmoid}(c_k) \text{ and } \tilde{w}_{k} = \text{sigmoid}(w_k)
\]

The resulting values become close to 1 when the frame is near the center of the moment, approximately 0.5 when it is at the start or end of the moment, and towards 0 as the frame is further away from the moment. The sharpness hyperparameter promotes a sharp contrast between the values inside and outside the moment. The value of the sharpness can be progressively increased with each iteration, enhancing the contrast over the course of the optimization (Figure~\ref{fig:sharpness}).

Since the soft moment mask is differentiable, it can be optimized in an end-to-end manner alongside the text generation module.
By introducing merely two parameters per moment, the optimization process of the temporal moment mask is both highly stable and efficient.
Moreover, our parameterization of a moment using center and width parameters provides straightforward interpretability and applicability.


\subsubsection{Pairwise temporal IoU loss}
\label{method:ptIou_loss}

To discover multiple temporal segments at the same time we optimize a group of moments for a video simultaneously, each with separate soft moment masking and prefix context. To encourage the model to capture different moments of distinct regions, we introduce pairwise temporal IoU loss between different moments. Pairwise temporal IoU loss between $N$ moments is calculated by the following equation:

\[
L_{\text{ptIoU}} = \frac{1}{{\binom{N}{2}}} \sum_{k=1}^{N-1} \sum_{l=k+1}^{N} \text{{IoU}}(m_k, m_l)
\]

In this expression, \(\binom{N}{2}\) represents the total number of possible pairwise combinations between $N$ moments. \(\text{{IoU}}(m_k, m_l)\) calculates the temporal Intersection over Union between the two moments $m_k$ and $m_l$.

\subsection{Joint optimization}
\label{method:joint_optimization}

The total loss of our method is the weighted sum of vision loss, language loss, and pairwise temporal IoU loss. The model is optimized in an end-to-end manner.

\[
L_{\text{total}} = \lambda_1 \cdot L_{\text{vision}} + \lambda_2 \cdot L_{\text{language}} + \lambda_3 \cdot L_{\text{ptIoU}}
\]

\(\lambda_1\), \(\lambda_2\), and \(\lambda_3\) are hyperparameters that represent the weights assigned to each loss term.

\section{Experiments}
\label{sec:results}

This section demonstrates the effectiveness of our proposed \ours model by comparing it to baselines and the state of the art. We begin by providing an overview of our experimental setup in Section \ref{subsec:setup}. We then present quantitative analysis in Section \ref{subsec:results}. Note that we add qualitative results in the Appendix Section \ref{sec:appendix:qualitative}.

\subsection{Experimental setup}
\label{subsec:setup}

\subsubsection{Datasets} 
\label{sec:setup:datasets}

For zero-shot dense video captioning, we use two datasets for evaluation: ActivityNet Captions~\citep{krishna2017dense} and YouCook2~\citep{zhou2018towards}. Adhering to a zero-shot setup, we refrained from using any caption or temporal annotations in training data.

\textbf{ActivityNet Captions} includes 20K untrimmed videos showcasing various human activities. Each video in this dataset lasts around 120 seconds on average and is annotated with an average of 3.7 temporally-localized captions.

\textbf{YouCook2} comprises 2K untrimmed cooking procedure videos, with an average duration of 320 seconds per video. Each video in the dataset is annotated with an average of 7.7 temporally-localized sentences.


\subsubsection{Implementation Details} 
We uniformly sample one frame per second from a given video. The visual feature extraction and text-image similarity calculation are done using the pre-trained CLIP ViT-L/14. We use the pretrained GPT-2 medium for the language model.

In the case of ActivityNet Captions, the number of moments $k$ for a video is set to 4. For the YouCook2 dataset, the number of moments $k$ for a video is set to 8. The initialization of the center and width parameters is based on the respective dataset distributions.

We set the vision loss weight to $\lambda_1 = 1$, the language loss weight to $\lambda_2 = 0.8$, and the pairwise temporal IoU loss weight to $\lambda_3 = 10$. The sharpness hyperparameter $\gamma$ is linearly increased starting from 10 and incremented by 1 after each generation iteration. The temperature hyperparameter $\tau$ is set to 1.0. Throughout the experiments, we employ 12 generation iterations. For the further implementation details, refer to \ref{sec:appendix:imple}

\subsubsection{Evaluation metrics}

For dense video captioning, we adopt three widely used metrics: CIDEr~\citep{vedantam2015cider} (C), METEOR~\citep{banerjee2005meteor} (M), and SODA\_c~\citep{fujita2020soda} (S). Both CIDEr and METEOR initially determine the matched pairs between the predicted moments and the ground truth annotations across IoU (Intersection over Union) thresholds of 0.3, 0.5, 0.7, and 0.9. The captioning metrics are then calculated based on these matched pairs. SODA\_c, on the other hand, addresses the limitations of traditional captioning metrics in the context of dense video captioning and considers the overarching narrative of the video.

\subsubsection{Baselines}

Since this work is the first attempt at zero-shot dense video captioning, there is no prior work directly addressing this task. Therefore, we evaluate our method by comparing it against several straightforward baseline approaches: 1) Scene detection using PySceneDetect\footnote{https://scenedetect.com} followed by image captioning with BLIP~\citep{li2022blip} (\textbf{PySceneDetect+BLIP}). PySceneDetect is a widely used scene detector for splitting a video into separate clips. We extract the center frame from each detected clip and use BLIP to generate corresponding captions. 2) Scene detection using PySceneDetect followed by a video captioner (\textbf{PySceneDetect+TimeSformer+GPT-2}). This is the same as the one with BLIP but uses an open-source pretrained video captioning model based on TimeSformer~\citep{bertasius2021space} and GPT2\footnote{https://huggingface.co/Neleac/timesformer-gpt2-video-captioning}. 3) Video captioning with TimeSformer+GPT2 model followed by frame matching with CLIP (\textbf{TimeSformer+GPT-2+CLIP}). This baseline first generates multiple captions using beam search with a video captioner and matches the most similar frame with each caption using CLIP. Then, the frame that best matches each caption is regarded as the center of the moment, with a fixed width applied across all moments. We add more implementation details of the baselines in the Appendix Section \ref{sec:appendix:baselines}.

\subsection{Results}
\label{subsec:results}

In this section, we evaluate and analyze the performance of our model in comparison to baselines and the current state-of-the-art models. Table \ref{table1} presents a performance comparison between our model, zero-shot baselines and methods that have stronger supervision. Table \ref{table2} shows a performance comparison in out-of-domain settings. We add more detailed ablation studies in the Appendix Section \ref{sec:appendix:ablation}.

\begin{table}[h]
\centering
\fontsize{8.0}{10}
\setlength{\tabcolsep}{3.0pt}
\renewcommand{\arraystretch}{1.3}
\begin{tabular}{lcccC{1.0cm}ccC{1.0cm}c}
\hline
\multirow{2}{*}{\textbf{Models}} & \multirow{2}{*}{\textbf{\begin{tabular}[c]{@{}c@{}}Trainable\\ Parameters\end{tabular}}} & \multirow{2}{*}{\textbf{Pretraining}} & \multicolumn{3}{c}{\textbf{ActivityNet Captions}} & \multicolumn{3}{c}{\textbf{YouCook2}} \\ \cline{4-9} &  &  & \textbf{S}  & \textbf{C}  & \textbf{M}  & \textbf{S}  & \textbf{C} & \textbf{M} \\ \hline
\multicolumn{9}{l}{\textit{\textbf{Full-training}}} \\ \hline
UEDVC \citep{zhang2022unifying} & 25M & \ding{55} & 5.5 & - & - & - & - & - \\
PDVC \citep{wang2021end} & 22M & \ding{55} & 6.0 & 29.3 & 7.6 & 4.9 & 28.9 & 5.7 \\
Vid2Seq (-) & 313M & \ding{55} & 5.4 & 18.8 & 7.1 & 4.0 & 18.0 & 4.6 \\
Vid2Seq & 313M & \checkmark & 5.8 & 30.1 & 8.5 & 7.9 & 47.1 & 9.3 \\ \hline
\multicolumn{9}{l}{\textit{\textbf{Few-Shot (1\%)}}} \\ \hline
Vid2Seq (-) & 313M & \ding{55} & 0.0 & 0.0 & 0.1 & 0.0 & 0.0 & 0.0 \\
Vid2Seq & 313M & \checkmark & 2.2 & 6.2 & 3.2 & 2.4 & 10.1 & 3.3 \\ \hline
\multicolumn{9}{l}{\textit{\textbf{Zero-Shot}}} \\ \hline
PySceneDetect+BLIP & - & \ding{55} & 1.1 & 2.7 & 1.0 & 0.5 & 1.6 & 0.6 \\
PySceneDetect+TimeSformer+GPT-2 & - & \ding{55} & 1.3 & 2.5 & 1.7 & 0.2 & 1.3 & 0.7 \\
TimeSformer+GPT-2+CLIP & - & \ding{55} & 1.6  & 3.1 & 2.1 & 0.7 & 1.9 & 0.8 \\
\ours~(ours) & 20M & \ding{55} & \normalsize\textbf{2.6} & \normalsize\textbf{7.5} & \normalsize\textbf{2.7} & \normalsize\textbf{1.6} & \normalsize\textbf{4.9} & \normalsize\textbf{2.1} \\ \hline
\end{tabular}
\vspace{0.3cm}
\caption{Performance comparison with other methods on the ActivityNet Captions and YouCook2 dataset across various models and supervision levels. \textit{Pretraining} column denotes whether the model is pretrained with video-text data. Vid2Seq (-) refers to the Vid2Seq model without pertaining. All results except for the zero-shot results are from corresponding papers. Best over zero-shot in bold.}
\label{table1}
\end{table}

\begin{table}[h]
\centering
\fontsize{8.0}{10}
\setlength{\tabcolsep}{7.0pt}
\renewcommand{\arraystretch}{1.5}
\begin{tabular}{l*{3}{>{\centering\arraybackslash}p{0.75cm}}*{3}{>{\centering\arraybackslash}p{0.75cm}}}

\hline
\multirow{2}{*}{\textbf{Models}} & \multicolumn{3}{c}{\textbf{ANet Captions} $\rightarrow$ \textbf{YouCook2}} & \multicolumn{3}{c}{\textbf{YouCook2} $\rightarrow$ \textbf{ANet Captions}} \\ 
\cline{2-7} & \textbf{S} & \textbf{C} & \textbf{M} & \textbf{S} & \textbf{C} & \textbf{M}  
\\ \hline
\multicolumn{7}{l}{\textit{\textbf{Full-training}}} \\ \hline
Vid2Seq & 0.02 & 0.1 & 0.03 & 0.2 & 0.5 & 0.2 \\ \hline
\multicolumn{7}{l}{\textit{\textbf{Zero-Shot}}} \\ \hline
\ours~(ours) & \normalsize\textbf{2.6} & \normalsize\textbf{7.5} & \normalsize\textbf{2.7} & \normalsize\textbf{1.6} & \normalsize\textbf{4.9} & \normalsize\textbf{2.1} \\ \hline
\end{tabular}

\vspace{0.3cm}
\caption{Comparison between our method and state-of-the-art fully-supervised method in out-of-domain settings. The results of Vid2Seq are from the official codebase and checkpoints. Best in bold.}
\label{table2}
\end{table}

\paragraph{Joint optimization is more effective than two-stage methods}
In dense video captioning, our model outperforms various zero-shot baselines on both ActivityNet Captions and YouCook2 datasets.
These baselines utilize two-stage approaches with a segmenting component and captioning component to tackle dense caption generation.
Despite the fact that the image captioning and the video captioning components of these baselines are trained directly using additional captioning data and captioning loss, there remains a noticeable gap in performance when compared to our approach.
This observation highlights the critical role of joint training in text generation and moment localization, enabling effective dense caption generation even in the absence of data.

\paragraph{\ours outperforms a state-of-the-art few-shot model}
Compared to models that have stronger supervision than ours, we observe that \ours surpasses the performance of few-shot Vid2Seq, a model with pretraining. It is worth noting that Vid2Seq is pretrained on the YT-Temporal-1B dataset, which consists of 18 million narrated videos spanning 1 billion frames paired with transcribed speech sentences. Remarkably, despite our model never having access to video data or temporal annotations, we achieved better performance than Vid2Seq fine-tuned with 1\% of the training data.

\paragraph{Text space of the target task and that of CLIP need to match}
YouCook2 shows a different trend compared to ActivityNet Captions. Here, our method underperforms the few-shot Vid2Seq. This divergence can be attributed to the distinct style of language annotation inherent to the dataset. ActivityNet Captions typically contain conventional captions briefly describing the visual content, such as "Cheerleaders are standing on the side of the road.". In contrast, YouCook2 is characterized by task-oriented, instructional textual annotations like "place a slice of cheese on the bread." Since our model relies on CLIP, which is pretrained with conventional image captions, the generated text resembles these captions. This style of resulting captions conflicts with YouCook2's ground truth captions, thus degrading performance in metrics. See Section~\ref{sec:limitation} for more discussion.

\paragraph{\ours is robust in out-of-domain setups}
Our method demonstrates greater robustness in out-of-domain setup, surpassing fully-trained state-of-the-art models. Unlike fine-tuned models, which are optimized for a target domain and thus struggle to adapt to new ones, our zero-shot approach maintains the performance across different domains. Its inherent domain-agnostic nature allows for flexibility, avoiding the overfitting pitfalls of specialized models.

\section{Limitation and Discussion}
\label{sec:limitation}
Our zero-shot method, by design, doesn't encounter any text or temporal annotation associated with the dataset. Consequently, it doesn't have the opportunity to learn the particular style of the output text and moments of the dataset. While this limitation could potentially be addressed by extending the method in various ways, including few-shot learning, we reserve this for future work.

\section{Conclusion}

In this work, we present a novel zero-shot method for dense video captioning, \ours, which utilizes soft moment masking and pairwise temporal IoU loss for end-to-end temporal localization. Our method, despite not requiring any video or annotations for training, not only surpasses various zero-shot baselines but also outperforms the state-of-the-art few-shot method on the widely-used benchmark, ActivityNet Captions. Moreover, it demonstrates superior robustness in out-of-domain scenarios compared to fully-supervised models, thereby showcasing its adaptability to diverse and previously unseen video data. 
This research not only presents a pioneering approach to zero-shot dense video captioning but also sheds light on the potential of aligning language and vision models. By combining the power of pretrained models of different modality, we can unlock new capabilities such as understanding temporal aspects in videos. These contributions advance the field of dense video captioning and offer valuable insights for future research in zero-shot alignment of language and vision models.

\bibliography{references}
\bibliographystyle{plainnat}

\clearpage
\appendix
\input{appendix}

\end{document}

%% file: appendix.tex
\section{Experimental setup}

In this section, we complement the description of our experimental setup outlined in Section \ref{subsec:setup}. We provide the implementation details (Section \ref{sec:appendix:imple}) and also give additional information about baselines (Section \ref{sec:appendix:baselines}).

\subsection{Implementation details}
\label{sec:appendix:imple}

We uniformly sample one frame per second from a given video. The visual features of each frame are extracted and the similarity between text and frames is measured using the pre-trained CLIP ViT-L/14. We use the pretrained GPT-2 medium for the language model.

For prefix context, we employ soft prompt of length 5. Also, we use the projected video embedding of length 20, i.e., we project the averaged frame CLIP embeddings to 20 token embeddings of GPT-2.

The initialization of the center and width parameters is based on the respective dataset distributions. In the case of ActivityNet Captions, the number of moments for a video $k$ is set to 4. The center parameters are initialized in a way that the sigmoid of their values uniformly transition from the start to the end of the video. The width parameter of each moment is initialized to -0.8472, resulting in a sigmoid value of 0.3.

For the YouCook2 dataset, the number of moments for a video $k$ is set to 8. The center parameters are initialized in a way that the sigmoid of their values uniformly transition from 0.1 to 0.9 of the video duration. This initialization aims to exclude irrelevant start and end frames, which usually contain intro and outro scenes. The width paramter of each moment is initialized to -2.1972, yielding a sigmoid value of 0.1. Additionally, a maximum width parameter value of -0.8472 is applied, which corresponds to 0.3 of the video duration.

We set the vision loss weight to $\lambda_1 = 1$, the language loss weight to $\lambda_2 = 0.8$, and the pairwise temporal IoU loss weight to $\lambda_3 = 10$. The temperature hyperparameter $\tau$ is set to 1.0. Throughout the experiments, we employ 12 generation iterations. The sharpness hyperparameter $\gamma$ is linearly increased starting from 10 and incremented by 1 after each generation iteration. For the generation of each new sentence, a hard prompt is randomly selected from the set of \{"Video showing", "Video shows", "Video of", "Photo showing", "Photo shows", "Photo of", "Picture showing", "Picture shows", "Picture of", "Image showing", "Image shows", "Image of"\}

The AdamW optimizer is employed with $\beta = (0.9, 0.999)$ and $\text{weight decay} = 0.0018$. We use a learning rate of $6e^{-3}$ with a cosine annealing learning rate scheduler. The experiments are conducted on NVIDIA A100 GPUs.

\subsection{Baselines}
\label{sec:appendix:baselines}

\paragraph{PySceneDetect+BLIP.} 
We split a given video using the default adaptive content detector method that analyzes the changes in average frame intensity/brightness using PySceneDetect. For image captioning, we use the BLIP Base image captioning model based on ViT-B/32. During decoding, we employ beam search with a beam size of 5 for caption generation.

\paragraph{PySceneDetect+TimeSformer+GPT-2.}
Here the configuration of PySceneDetect is the same as PySceneDetect+BLIP. For video captioning, we employ an open-source pretrained video captioner based on TimeSformer and GPT2. We use beam search with a beam size of 8 for decoding.

\paragraph{TimeSformer+GPT-2+CLIP.}
In this baseline, instead of initially splitting the video, we first perform the captioning process, following which each caption is matched to a specific moment within the video. To generate multiple captions from a video, we use a beam search with a beam size of 8. We employ the same video captioner as PySceneDetect+VideoCaptioner. Subsequently, we compute the CLIP scores to measure the similarity between each generated caption and all the frames within the video. We use CLIP ViT-B/32 for this process. The frame with the highest CLIP score is considered the central frame for the moment associated with that particular caption. Finally, we apply a fixed width of 0.3 of the total duration to each moment.

\section{Qualitative results}
\label{sec:appendix:qualitative}

\begin{figure}[h]
\centering
\includegraphics[width=120mm]{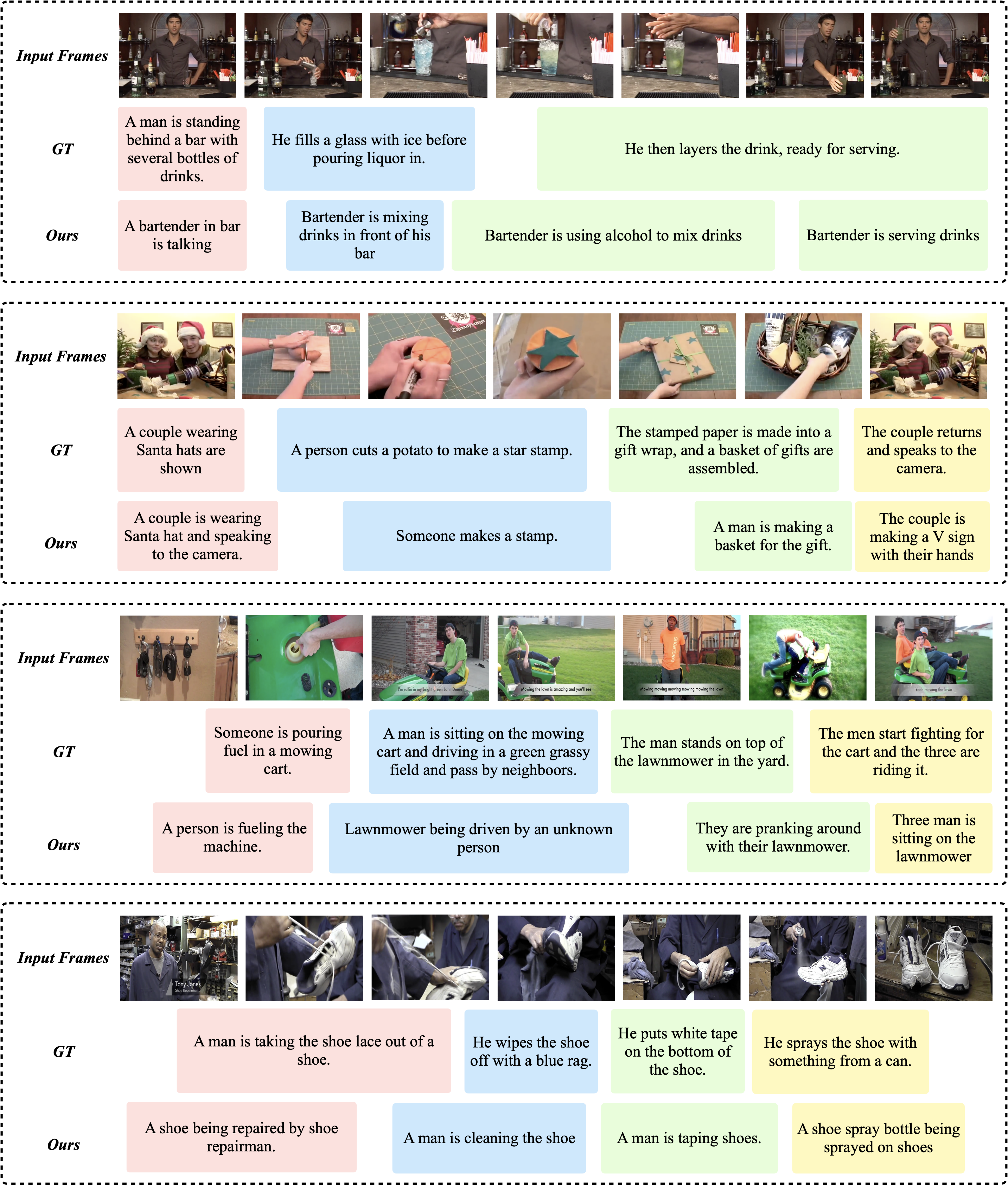}
\caption{Examples of dense video captioning predictions generated by \ours on the validation set of ActivityNet Captions, along with the ground-truth annotations.}
\label{fig:appendix:qualitative:anet}
\end{figure}

\begin{figure}[h]
\centering
\includegraphics[width=120mm]{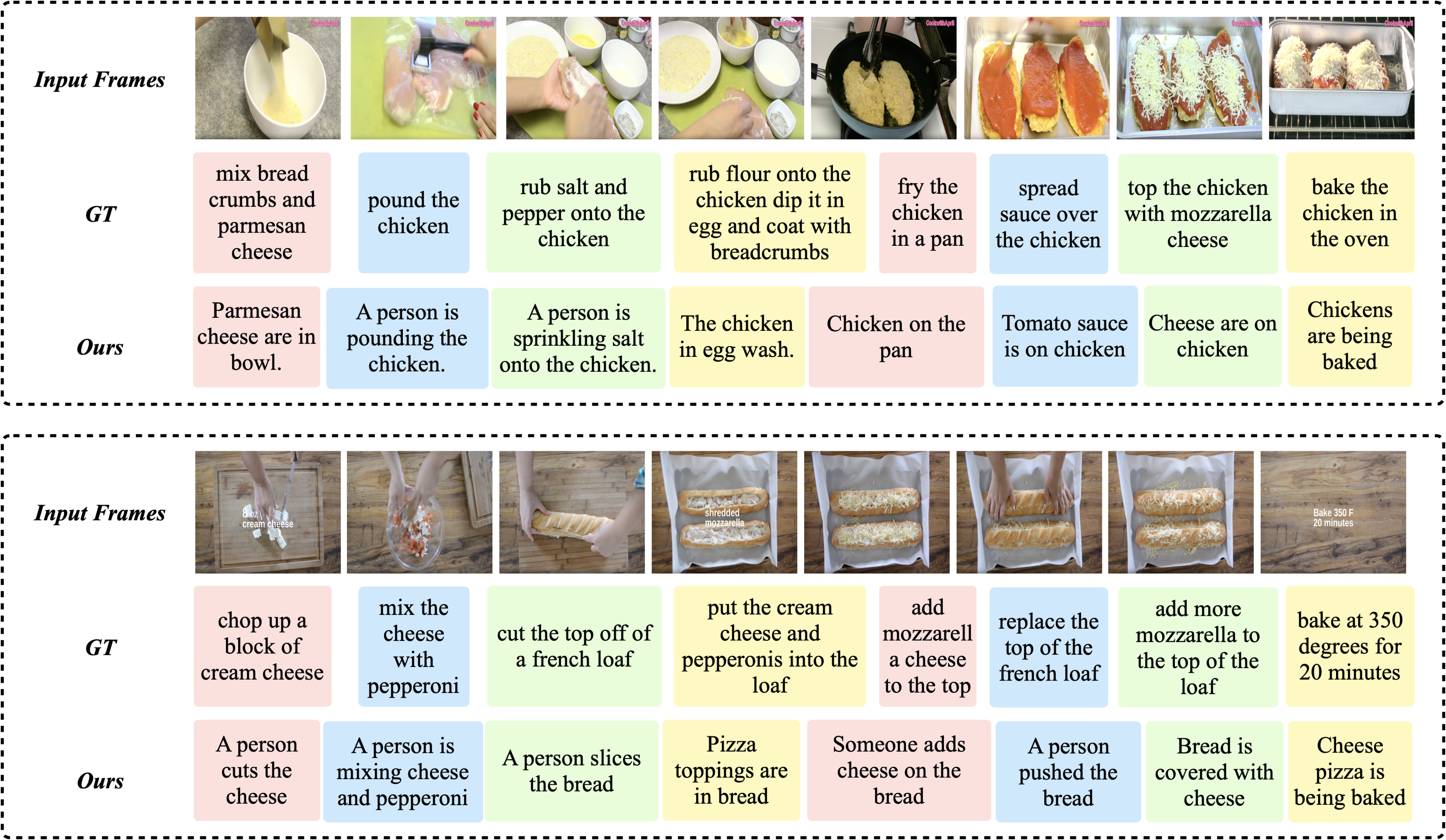}
\caption{Examples of dense video captioning predictions generated by \ours on the validation set of YouCook2, along with the ground-truth annotations.}
\label{fig:appendix:qualitative:youcook2}
\end{figure}

\begin{figure}[h]
\centering
\includegraphics[width=120mm]{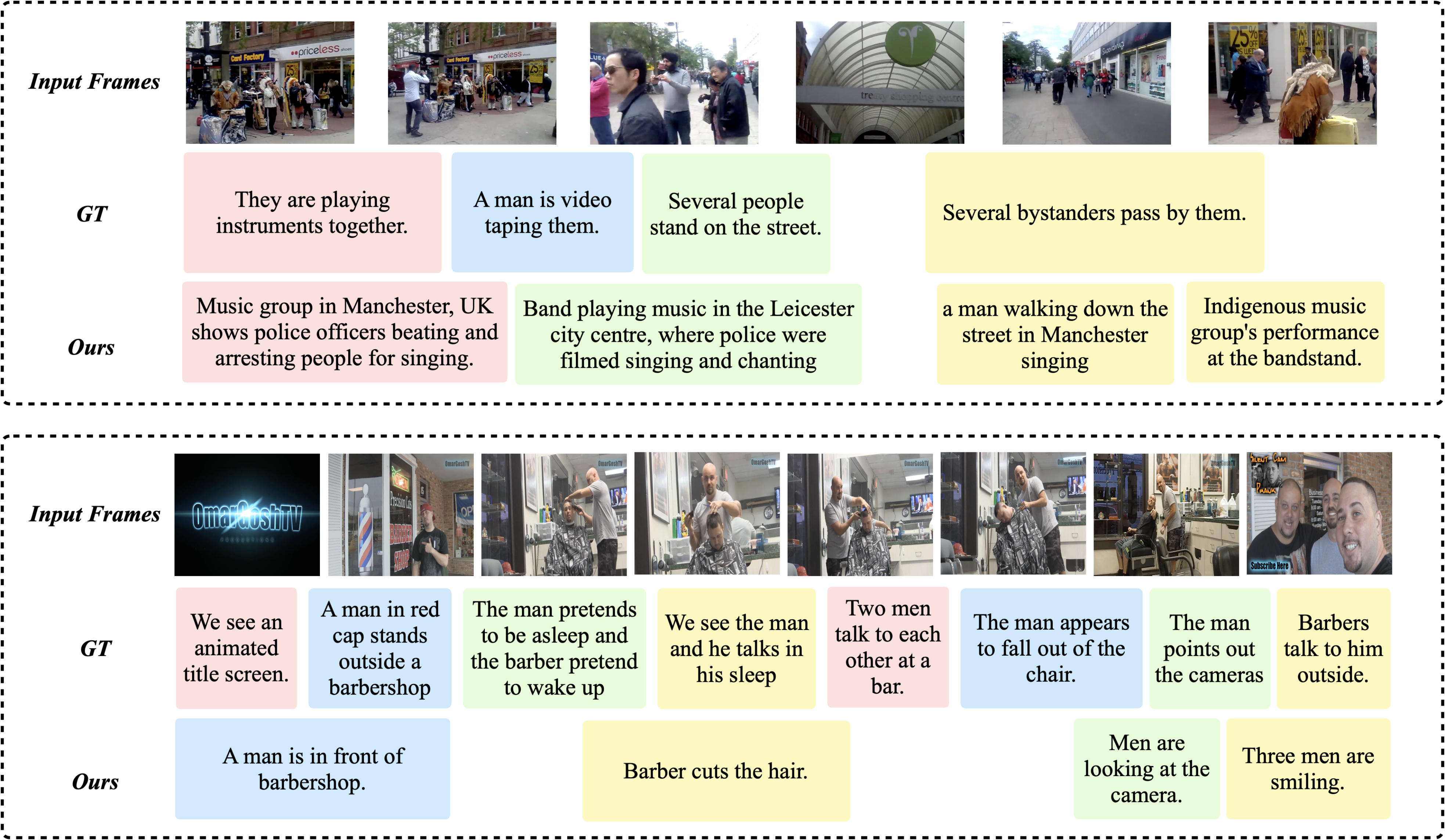}
\caption{Failure cases of \ours on the validation set of ActivityNet Captions. These examples show situations where the model (1) generates captions lacking visual evidence, and (2) is unable to capture all significant moments due to the fixed number of moments per video.}
\label{fig:appendix:qualitative:fail}
\end{figure}

Figure \ref{fig:example} presents the qualitative results of dense event captioning obtained by our \ours model. Here, We show additional results attained from the ActivityNet Captions and YouCook2 datasets in Figures \ref{fig:appendix:qualitative:anet}, \ref{fig:appendix:qualitative:youcook2}, and \ref{fig:appendix:qualitative:fail}.

Figure \ref{fig:appendix:qualitative:anet} shows that \ours can capture meaningful moments and generate corresponding captions, even without any training data. In Figure \ref{fig:appendix:qualitative:youcook2}, we observe that although the style of the caption may differ from the ground truth (as discussed in Section \ref{sec:limitation}), \ours still manages to generate meaningful dense captions and identify moment boundaries.

Figure \ref{fig:appendix:qualitative:fail} illustrates failure cases, such as (1) generating descriptions that lack visual grounding (i.e., hallucination) and (2) failing to capture all significant moments due to the fixed number of moments per video.

\section{Ablation studies}
\label{sec:appendix:ablation}

In this section, we provide ablation studies that complement the results presented in Section \ref{subsec:results}. We use the same default hyperparameters, evaluation metrics, and downstream datasets for these experiments.

\paragraph{Vision-language similarity model}

In Table \ref{table:appendix:clip}, we analyze the benefits of scaling up the size of the pretrained CLIP model. We find that scaling up the CLIP size from ViT-B/32 to ViT-L/14 brings considerable performance improvements. These results suggest that further performance improvements could potentially be achieved by scaling up CLIP to even larger models. Due to computational constraints, we did not conduct experiments with CLIP models larger than ViT-L/14, leaving this as an area for future exploration.

\begin{table}[h]
\centering
\setlength{\tabcolsep}{5.0pt}
\renewcommand{\arraystretch}{1.5}

\begin{tabular}{lcccccc}
\hline
\multirow{2}{*}{\textbf{Models}} & \multicolumn{3}{c}{\textbf{ActivityNet Captions}} & \multicolumn{3}{c}{\textbf{YouCook2}} \\ \cline{2-7} & \textbf{SODA\_c} & \textbf{CIDEr} & \textbf{METEOR} & \textbf{SODA\_c} & \textbf{CIDEr} & \textbf{METEOR} \\ \hline
\textbf{ViT-B/32} & 2.4 & 6.6 & 2.4 & 1.5 & 4.5 & 1.8 \\
\textbf{ViT-L/14} & \textbf{2.6} & \textbf{7.5} & \textbf{2.7} & \textbf{1.6} & \textbf{4.9} & \textbf{2.1} \\ \hline
\end{tabular}

\vspace{0.3cm}
\caption{Vision-language similarity model ablation.}
\label{table:appendix:clip}
\end{table}

\paragraph{Language model}

In Table \ref{table:appendix:lm}, we evaluate the effect of scaling up the size of the pretrained GPT-2 language model. We find that scaling up the language model size also increases the overall performance of the model. Note that, due to computational constraints, our default setting across all other experiments is GPT-2 medium.

\begin{table}[h]
\centering
\setlength{\tabcolsep}{5.0pt}
\renewcommand{\arraystretch}{1.5}

\begin{tabular}{lcccccc}
\hline
\multirow{2}{*}{\textbf{Models}} & \multicolumn{3}{c}{\textbf{ActivityNet Captions}} & \multicolumn{3}{c}{\textbf{YouCook2}} \\ \cline{2-7} & \textbf{SODA\_c} & \textbf{CIDEr} & \textbf{METEOR} & \textbf{SODA\_c} & \textbf{CIDEr} & \textbf{METEOR} \\ \hline
\textbf{GPT-2 medium} & 2.6 & 7.5 & 2.7 & 1.6 & 4.9 & 2.1 \\
\textbf{GPT-2 large}  & \textbf{2.6} & \textbf{7.6} & \textbf{2.8} & \textbf{1.8} & \textbf{5.2} & \textbf{2.3} \\ \hline
\end{tabular}

\vspace{0.3cm}
\caption{Language model ablation.}
\label{table:appendix:lm}
\end{table}

\paragraph{Projected video embedding}

Table \ref{table:appendix:projected} presents an ablation study on the projected video embedding part of the prefix context presented in Section \ref{method: text_generation}. By default, the process projects the averaged video frame embeddings into 20 token embeddings. We find that incorporating the projected video embeddings in the prefix context results in improved performance compared to the model without it. Additionally, projecting video embeddings into a greater number of tokens is beneficial.



\begin{table}[h]
\centering
\setlength{\tabcolsep}{5.0pt}
\renewcommand{\arraystretch}{1.5}

\begin{tabular}{cccc}
\hline
\multirow{2}{*}{\textbf{\begin{tabular}[c]{@{}c@{}}\# of projected embeddings\end{tabular}}} & \multicolumn{3}{c}{\textbf{ActivityNet Captions}} \\ \cline{2-4} & \textbf{SODA\_c} & \textbf{CIDEr} & \textbf{METEOR} \\ \hline
\textbf{No projected embeddings} & 2.4 & 7.1 & 2.6 \\
\textbf{10} & 2.5 & 7.1 & 2.6 \\
\textbf{20} & \textbf{2.6} & \textbf{7.5} & \textbf{2.7} \\ \hline
\end{tabular}

\vspace{0.3cm}
\caption{Projected video embedding ablation.}
\label{table:appendix:projected}
\end{table}

\paragraph{Sharpness of soft moment mask}
In the default settings, we increment the sharpness hyperparameter $\gamma$ by 1 after each generation iteration, starting from a base value of 10. In Table \ref{table:appendix:sharpness}, we ablate the scheduling of the sharpness hyperparameter. Our findings suggest that starting with a lower sharpness value and gradually increasing it can lead to better performance than constant scheduling schemes.

\begin{table}[h]
\centering
\setlength{\tabcolsep}{5.0pt}
\renewcommand{\arraystretch}{1.5}

\begin{tabular}{lccc}
\hline
\multirow{2}{*}{\textbf{Scheduling}} & \multicolumn{3}{c}{\textbf{ActivityNet Captions}}   \\ \cline{2-4} & \textbf{SODA\_c} & \textbf{CIDEr} & \textbf{METEOR} \\ \hline
\textbf{Linear (initial \(\gamma\)=10)*} & \textbf{2.6} & \textbf{7.5} & \textbf{2.7} \\
\textbf{Constant (\(\gamma\)=10)} & 2.5 & 7.1 & 2.6 \\
\textbf{Constant (\(\gamma\)=20)} & 2.4 & 6.8 & 2.4 \\ \hline
\end{tabular}

\vspace{0.3cm}
\caption{Sharpness of soft moment mask ablation. Linear (initial $\gamma=10$)* linearly increment $\gamma$ by 1 after each iteration, starting from $\gamma=10$.}
\label{table:appendix:sharpness}
\end{table}
